\documentclass[letterpaper]{article} 
\usepackage[preprint]{aaai2027}  

\usepackage{times}
\usepackage{latexsym}

\usepackage[T1]{fontenc}

\usepackage[hyphens]{url}  
\usepackage{graphicx} 
\usepackage{natbib}  
\usepackage{caption} 
\usepackage[utf8]{inputenc}

\usepackage{microtype}

\usepackage{inconsolata}
\usepackage{newfloat}

\usepackage{amssymb}
\usepackage{amsmath,bm}
\usepackage{mathrsfs} 
\usepackage{amsthm}%
\usepackage{mathtools}
\usepackage{mathrsfs}%
\usepackage{algorithm}
\usepackage{algpseudocode}
\usepackage{booktabs}       
\usepackage{lineno}
\usepackage{amsfonts}       

\usepackage{nicefrac}       
\usepackage{xcolor}         
\usepackage{multirow}
\usepackage{subfigure}
\usepackage{tcolorbox}

\usepackage{listings}
\DeclareCaptionStyle{ruled}{labelfont=normalfont,labelsep=colon,strut=off} 
\floatstyle{ruled}
\newfloat{listing}{tb}{lst}{}
\floatname{listing}{Listing}

\title{NeSyFS: A Neuro-symbolic Fast-Slow Thinking Framework for LLM Agent under Partial Observability}
\author{
    Duo Xu, Faramarz Fekri\thanks{This work is supported by ARO grant under Award \#W911NF-23-1-0146 and a gift by Open Philanthropy.}
}
\affiliations{
    Georgia Institute of Technology\\
    225 North Ave NW, Atlanta, GA 30332\\
    dxu301@gatech.edu
}

\begin{document}

\maketitle

\begin{abstract}
Recently Large Language Models (LLMs) have been increasingly deployed as autonomous agents in applications such as self-reflection, retrieval-augmented generation, and scientific discovery. In these settings, agents must act based on limited observations rather than full environmental states, leading to partial observability. This introduces several key challenges: belief state inference, task objective misalignment, and planning under uncertainty.
Prior approaches typically condition actions on full or summarized action-observation histories whose redundant and irrelevant information can mislead the decision making of LLM agent.
Inspired by human cognition, we propose a novel \textbf{Ne}uro-\textbf{Sy}mbolic \textbf{F}ast-\textbf{S}low thinking (NeSyFS) framework for LLM agent, addressing the challenges introduced by partial observability in a unified approach. We use a knowledge graph (KG) to represent the belief state, providing triplets as context for every module of NeSyFS. The fast-thinking module performs reactive action, while slow-thinking conducts a new uncertainty-aware planning by following the high-level structure of twisted sequential Monte Carlo (TSMC) algorithm. To mitigate the misalignment of task objective, a reflection module is used to reflect fast-thinking actions, and also switches to the slow-thinking module whenever reactive actions repeatedly fail. Experiments on three representative benchmarks, i.e. ALFWorld, Webshop, and ScienceWorld, demonstrate significant advantages over previous methods.
\end{abstract}

\begin{figure}
    \centering
    \includegraphics[width=3in]{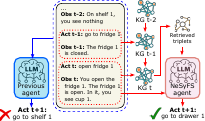}
    \caption{Comparison of previous and NeSyFS agent. Previous agent uses full or summarized interaction history as context for decision making, while our agent updates a KG-based memory dynamically and uses KG-retrieved triplets as context. The KG $\mathcal{G}_{t-1}$ is updated with action $a_t$ and observation $o_t$, i.e. $\mathcal{G}_t=\text{Update}(\mathcal{G}_{t-1}, o_t, a_t)$. The triplets related with solving task $\eta$ are retrieved from KG, i.e. $\tau_t=\mathcal{R}(\mathcal{G}_t,\eta)$, which can provide structured context to help LLM make correct decision.}
    \label{fig:nesyfs_intro}
\end{figure}

\section{Introduction}
Large Language Models (LLMs) have achieved notable success in agentic applications by extending beyond static text generation to support autonomous decision-making, enabling capabilities such as tool use \citep{schick2023toolformer}, retrieval-augmented reasoning \citep{gao2023retrieval}, multi-step planning \citep{rawat2025pre}, and self-reflection \citep{renze2024self} in complex environments. These advances have allowed LLM agents to perform tasks in domains ranging from embodied interaction to software automation \citep{jin2024llms} and scientific workflows \citep{yamada2025ai}. However, a fundamental challenge arises from partial observability: unlike supervised settings with complete input context, agentic environments typically reveal only local and immediate observations, while the true underlying state remains latent \citep{shridhar2020alfworld,yao2022webshop}. As a result, LLM agents must infer a belief over the hidden state from incomplete and potentially noisy interaction histories \citep{xi2024agentgym,ma2024agentboard}, which are often long and difficult to compress effectively within limited context windows. In addition, since the agent must act without access to the true underlying state, increased amount of uncertainty is introduced, and planning under uncertainty becomes a challenge for the LLM agent. Moreover, the uncertainty and missing state information from partial observability can make the LLM agent produce actions misaligned with task objective, degrading the performance of task completion.

In this work, we propose a novel neuro-symbolic fast–slow thinking framework (NeSyFS) for LLM agents that addresses the challenges of partial observability in a unified manner, enabling latent state inference, mitigating task-objective misalignment, and supporting planning under uncertainty. Unlike prior full-history-based or summarization-based approaches that rely on unstructured memory, we employ a knowledge graph (KG) to represent the latent state of the environment, thereby encoding interaction histories into a structured and symbolic memory. Specifically, each incoming observation is transformed into triplets that capture relationships and attributes of entities and events. Newly acquired facts are incorporated into the KG, while outdated triplets in KG are pruned, ensuring a consistent and dynamically updated representation of the latent state. The KG-retrieved triplets are used as context in every module of NeSyFS.

Furthermore, inspired by human cognition \citep{kahneman2011thinking}, the proposed framework decomposes decision-making into fast and slow thinking processes. The fast thinking module generates reactive actions by retrieving relevant triplets from the KG to construct a symbolic approximation of the latent state. In contrast, the slow thinking module performs deliberative planning under uncertainty. Slow thinking is used only when fast-thinking decision making has too much uncertainty or is unreliable. To this end, we propose a new uncertainty-aware planning algorithm which follows the high-level structure of twisted sequential Monte Carlo (TSMC) \citep{del2006sequential}. Specifically, it maintains a set of particles and uses resampling to guide them toward the target distribution modeling task-completing trajectories. LLM is prompted to predict next observation and existence of task progress in the state transition. The sampling mechanism of this method can make the particles robust to mistakes in LLM-based evaluations of task progress, and reach a good balance between exploration and exploitation. As far as we know, we are the first to develop a TSMC-style planning method for LLM agent. 

Although fast thinking emphasizes locally plausible actions, it does not explicitly guarantee alignment with global task objectives. To address this limitation, we incorporate a reflection module that evaluates the consistency of fast-thinking actions with the task objective, conditioned on relevant triplets retrieved from the KG. To control computational overhead, slow thinking is invoked only when fast-thinking actions repeatedly fail in this reflective evaluation. With KG-provided triplets as context, stepwise reflection is enabled and is more efficient and robust than conventional trajectory-wise reflection.

Our contributions are summarized as follows.
\begin{itemize}
    \item We propose a neuro-symbolic fast-slow thinking framework that addresses challenges introduced by partial observability in a unifying manner.
    \item Based on the context retrieved from the KG, a neuro-symbolic TSMC-style planning algorithm is introduced within the slow-thinking module to address the uncertainty under partial observability.
    \item We design a KG-augmented reflection module that aligns fast-thinking actions with task objectives. Using KG-retrieved triplets, the module enables stepwise reflection that is more efficient and robust than conventional trajectory-level reflection.
    \item Experiments on three benchmarks demonstrate that using KG-retrieved triplets as context improves the LLM performance on decision-making, reflection, and world modeling. 
\end{itemize}


\section{Preliminary Background}
\label{sec:background}

\subsection{Partial Observability}
\label{sec:pomdp}
We consider agentic tasks in which an autonomous LLM responds to a user query through iterative interaction with an external environment $\mathcal{E}$. Each episode begins with a user query and unfolds over a finite horizon $H$. 

Such agentic tasks can be naturally formulated as a Partially Observable Markov Decision Process (POMDP) \citep{he2024words,zhang2025landscape}, represented by the tuple $\mathcal{M}=(\mathcal{S},\mathcal{A},\mathcal{O},\mathcal{T},O,R)$. Here, $\mathcal{S}$ denotes the latent state space of the environment, $\mathcal{A}$ the action space, and $\mathcal{O}$ the observation space consisting of textual environmental response. The environment dynamics are governed by the transition function $\mathcal{T}(s' \mid s,a)$ and the observation model $O(o \mid s,a)$. The reward function $R(s)$ is defined only on terminal states and indicates task success.

At time step $t$, the agent executes an action $a_t$, after which the environment transitions to a latent state $s_t \sim \mathcal{T}(\cdot \mid s_{t-1}, a_{t})$ and returns an observation $o_t \sim O(\cdot \mid s_t, a_t)$. During inference, neither the latent state $s_t$ nor the reward function is directly observable to the agent. This results in partial observability, which introduces several fundamental challenges, including latent state inference, task objective misalignment, and planning under uncertainty. In particular, inferring $s_t$ from the interaction history is difficult because the history often contains substantial redundant and noisy information. Moreover, not knowing $s_t$ creates a lot of uncertainty and further complicates the prediction of future situations in planning. In this work, we propose a novel framework to unifiedly resolve these challenges.

\subsection{Knowledge Graph}
\label{sec:knowledge_graph}
Knowledge graphs (KGs) \citep{ehrlinger2016towards} represent information as graph-structured data, where nodes correspond to entities and edges encode relationships among them. KGs can be constructed from diverse sources, including structured databases, unstructured text, and other heterogeneous data modalities. Similar to traditional databases, graph databases support structured querying mechanisms, such as the Cypher query language \citep{francis2018cypher}.

The relationship between a pair of nodes in a KG is commonly referred to as a triplet of subject, predicate, and object. In this work, we consider the following triplet forms:
\begin{itemize}
    \item (entity 1, relationship, entity 2)
    \item (entity 1, attribute, boolean/value)
    \item (entity 1, verb in the past tense, entity 2)
\end{itemize}
which describe relationships, attributes and historic events of entities. A KG is denoted as $\mathcal{G}=(V,E)$ where $V$ is a set of semantic vertices corresponding to entities in the environment, and $E$ is a set of semantic edges representing triplets in the above forms.
Our framework has a retrieval-augmented generation (RAG) component which retrieves the relevant triplets from a KG and feeds them to LLM as context approximating the underlying state of the POMDP environment.


\subsection{Twisted Sequential Monte Carlo}
\label{sec:tsmc}

Twisted Sequential Monte Carlo (TSMC) \citep{doucet2001introduction,del2006sequential,chopin2020introduction,zhao2024probabilistic,feng2025step} is a probabilistic inference framework that incrementally guides particles toward high-probability regions of a target distribution $\sigma$ through intermediate twisted distributions $\psi_t$. At each step, particles are propagated using a proposal distribution and resampled according to importance weights, allowing computational effort to focus on promising trajectories while reducing variance. In this work, to address uncertainty under partial observability, the slow-thinking module follows a TSMC-like structure, where LLMs are used to propose future states and observations and to score particles. 

Sequential Monte Carlo (SMC) and its variants have been applied into planning and reinforcement learning by many previous work \citep{piche2018probabilistic,macfarlane2024spo,abdulsamad2026sequential}. Compared with other planning methods, such as Monte Carlo Tree Search (MCTS) \citep{chen2024tree}, SMC-style methods can better address the uncertainty of states and reduce computational complexity, since particles representing less-likely states will be discarded and not branched in large probability. However, in tree-search-based methods, every state will be branched and its child nodes will be evaluated. In this work, we develop a TSMC-style LLM-based planning method for slow-thinking module, where the state is approximated by related triplets retrieved from memory KG. As far as we know, this work is the first of doing so.

\begin{figure}
    \centering
    \includegraphics[width=3.in]{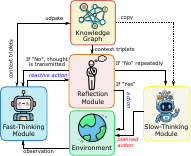}
    \caption{\small Diagram of NeSyFS framework. Knowledge graph (KG), representing the latent state of environment, is updated by fast-thinking module with new observation and retrieved by all the modules. Every reactive action is reflected. If "No", fast-thinking is called again with thought from reflection module. If "No" repeatedly for $K$ number of times, the slow-thinking module is initiated to search optimal action via planning, where memory KG is copied to initialize the simulation of future latent states and observations. 
    }
    \label{fig:proposed_framework}
\end{figure}

\section{Methodology}
\label{sec:method}
In this section, we introduce the NeSyFS framework, which maintains a  memory KG as a proxy for the latent state of the environment. The diagram of the proposed framework is shown in Figure \ref{fig:proposed_framework}. Using triplets retrieved from the memory KG as a context, the fast-thinking module generates reactive actions, while the slow-thinking module performs uncertainty-aware deliberative planning. A reflection module evaluates the fast-thinking actions with respect to the task objective, where the KG-retrieved triplets facilitate reflection in the stepwise granularity. We will first introduce the fast-thinking module, including the memory KG. Then, reflection module will be presented with details. Finally, the slow-thinking module, including the proposed TSMC-style planning algorithm, is introduced.

\subsection{Fast Thinking Module}
\label{sec:fast}

We use a KG $\mathcal{G}=(V,E)$ as working memory to represent the latent state of the environment (POMDP $\mathcal{M}$). Instead of using the entire interaction history or its summary as unstructured context, this memory organizes environmental knowledge into structured triplets which describe attributes and relations of entities and past events, in the forms in Section \ref{sec:knowledge_graph}. 

\begin{figure}
    \centering
    \includegraphics[width=2.2in]{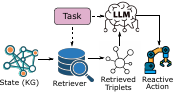}
    \caption{Diagram of decision making in fast-thinking module.}
    \label{fig:thrust1_diagram_policy}
\end{figure}

At each time step $t$, given the current observation $o_t$ and task description $\eta$, the fast-thinking module first retrieves task-relevant triplets $\tau_t$ from memory KG $\mathcal{G}_t$ by calling the retrieval method $\mathcal{R}(\mathcal{G}_t,\eta)$. The details of $\mathcal{R}$ are introduced in Appendix \ref{sec:retrieval_KG}. Then, $\tau_t$ are then provided as contextual input to the LLM which is prompted to produce the next action for accomplishing task $\eta$ with observation $o_t$. During this process, chain-of-thought (CoT) reasoning \citep{wei2022chain} is employed, making the fast-thinking action selection process analogous to the ReAct method \citep{yao2022react}. This can be formally expressed as $a_t=LLM(P^{\text{fast}};\mathcal{R}(\mathcal{G}_t),o_t,\eta)$, where $P^{\text{fast}}$ is the reactive decision making prompt template, $\mathcal{G}_t$ is the KG representing current latent state. Diagram of decision making in the fast-thinking module is shown in Figure \ref{fig:thrust1_diagram_policy}.

Whenever $a_t$ is applied into the environment, the agent receives a new observation $o_t$. The KG $\mathcal{G}_t$ will be updated with $o_t$ and $a_t$ to produce $\mathcal{G}_{t+1}$ to represent next state. This process is denoted as $\mathcal{G}_{t+1}=\text{Update}(\mathcal{G}_t,o_t,a_t)$, where new triplets from $o_t$ and $a_t$ will be added to $\mathcal{G}_t$, and outdated triplets in $\mathcal{G}_t$ will be removed. The details of update process are presented in Appendix \ref{sec:update_KG}. 

\begin{figure*}
    \centering
    \includegraphics[width=6.1in]{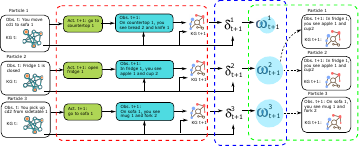}
    \caption{Diagram of the proposed TSMC-style uncertainty-aware planning algorithm, where three particles are updated in the planning step $t$, consisting of propagation, weight update and resampling processes. The red box shows the propagation process, where the prediction of KG at time $t+1$ is essentially updating KG at $t$ with $a_{t+1}$ and $o_{t+1}$. The blue box is the weight update process, where the blue circle represents the value of an updated weight. The green box represents the process of resampling. Every particle stores all the historic observations, actions, states (KG), and weights.}
    \label{fig:tsmc_planning}
\end{figure*}

\subsection{Reflection Module}
\label{sec:reflection}
Due to incomplete information about the underlying environment state, partial observability can introduce misalignment between the agent’s actions and the task objective, which has emerged as a major challenge for LLM agents \citep{fang2025preemptive,kim2025reflact,chung2025evaluating}. To mitigate this issue, we introduce a prompt-based reflection module that detects misalignment between the actions produced by the fast-thinking component and the task objective. Unlike self-reflection methods \citep{shinn2023reflexion}, in which the agent introspects and explains its own behavior, the proposed reflection module adopts a third-person perspective to verify whether a candidate action aligns with the task objective. Our innovation here is to use the KG-retrieved triplets as context to facilitate reflection in the stepwise granularity, since context triplets provide sufficient state information for reflection. 

The reflection can be expressed as $Y^{\text{ref}}=LLM(P^{\text{ref}};a_t, o_t,\mathcal{R}(\mathcal{G}_t),\eta)$, where $P^{\text{ref}}$ is the prompt for refection, $a_t$ is the fast-thinking action, $\mathcal{R}$ is the operation of KG retrieval introduced in previous section, $\eta$ is the task description, and output $Y^{\text{ref}}\in\{\text{Yes},\text{No}\}$ which indicates whether $a_t$ aligns with $\eta$ or not. 
To mitigate the randomness of LLM output, we propose to adopt a self-consistency method \citep{wang2022self} which draws $n$ samples of $Y^{\text{ref}}$ from LLM and sets the final output $\bar{Y}^{\text{ref}}$ as "Yes" only when $80\%$ of $Y^{\text{ref}}$ samples are "Yes". We use $n=5$ in experiments.

As shown in Figure \ref{fig:proposed_framework}, if the final reflection answer $\bar{Y}^{\text{ref}}$ is "Yes", then $a_t$ is applied to the environment. Otherwise, the thought of reflection module which produces "No" is sent back to the fast-thinking module and a new reactive action is produced by calling the fast-thinking module again. If $\bar{Y}^{\text{ref}}$ is repeatedly not "Yes" for $K$ number of times, say $K=3$, the slow thinking module is activated.


\subsection{Slow Thinking Module}
\label{sec:slow_thinking}

As shown in Figure \ref{fig:proposed_framework}, when the fast-thinking is unreliable, the agent transitions to the slow-thinking module, which performs deliberative planning by explicitly searching over candidate action sequences in a step-by-step manner. To handle the uncertainty arising from partial observability, we propose a novel neuro-symbolic uncertainty-aware planning algorithm which follows the high-level framework of twisted sequential Monte Carlo (TSMC) algorithm introduced in Section \ref{sec:tsmc}. 

The proposed planning algorithm is illustrated in Figure \ref{fig:tsmc_planning}. Specifically, the algorithm maintains a set of weighted particles that represent possible future observations, latent states (i.e., KGs), and actions over a planning horizon $H$. To identify an optimal plan, these particles are guided toward high-probability regions of a target distribution $\sigma$, which corresponds to trajectories that successfully accomplish the task. At each planning step, every particle is propagated by predicting the next action, observation, and latent state (KG). The particle weights are then updated according to the likelihood of the newly predicted state achieving task progress. Finally, a resampling procedure based on the updated weights preserves particles that make meaningful task progress while filtering out less promising particles with a large probability.

To address the difficulty LLMs face in evaluating task progress \citep{setlur2025rewarding,zheng2025survey}, particle weights are updated using binary LLM feedback indicating whether a given state transition makes task progress or not. Moreover, the concise state representation provided by KG can improve the correctness of LLM judgment. In addition, the resampling mechanism can make our approach more robust to LLM's mistakes, as it still allows promising particles to survive even when they are temporarily misjudged by LLM. 

Compared with previous tree-search planning methods, such as MCTS \citep{chen2024tree}, the proposed TSMC-style method is better at addressing state uncertainty and more computationally efficient. Previous methods are difficult at predicting next state or observation under partial observability, while the proposed method leverages memory KG to accurately predict next KG by updating current KG with new action and observation. In addition, in tree-search methods, every leaf node is expanded by predicting multiple next states as child nodes. However, in the proposed method, particles with poor task progress will be discarded and not be expanded with a large probability, hence reducing the computation complexity.

\noindent
{\bf Definition.}
Denote $N$ as number of particles. Denote particle $i$ at the planning step $\tau$ as $\bm{p}_{\tau}^i=(\hat{\bm{o}}^i_{1:\tau},\hat{\bm{\mathcal{G}}}^i_{1:\tau},\hat{\bm{a}}^i_{1:\tau},\omega^i_{\tau})$, where $\bm{o}^i_{1:\tau}=(\hat{o}^i_1,\ldots,\hat{o}^i_{\tau})$ is the sequence of observations predicted by LLM until time step $\tau$, and same for $\hat{\bm{\mathcal{G}}}^i_{1:\tau}$ and $\hat{\bm{a}}^i_{1:\tau}$. 

\noindent
{\bf Initialization.} Denote $t_0$ as the time step when slow thinking is activated. For each particle, set the first observation, latent state and action as copies of the actual observation, state (KG), and action at time step $t_0$, i.e. $\hat{o}^i_1=o_{t_0}, \hat{\mathcal{G}}^i_1=\mathcal{G}_{t_0}, \hat{a}^i_1=a_{t_0}$ for $i=1,\ldots,N$, so that every particle starts from the same origin. Set the weight of each particle to be 1, i.e. $\omega^i_1=1$.

\noindent
{\bf Propagation.} As shown in the red box of Figure \ref{fig:tsmc_planning}, in each planning step $\tau$, every particle $\bm{p}_{\tau}^i$ is propagated forward to generate a new particle $\bm{p}_{\tau+1}^i$ by predicting the next action $\hat{a}_{\tau+1}^i$, observation $\hat{o}_{\tau+1}^i$, and latent state $\hat{\mathcal{G}}_{\tau+1}^i$ one-by-one. 

First, the next action is produced by an LLM-based policy, described as $\hat{a}_{\tau+1}^i=LLM(P^{\text{fast}}; \mathcal{R}(\hat{\mathcal{G}}_{\tau}^i), \hat{o}_{\tau}^i, \eta)$, where $P^{\text{fast}}$ is the prompt template used in the fast-thinking module, and $\eta$ is the task description. 

Subsequently, for each particle $i$, we need to predict the next observation $\hat{o}^i_{\tau+1}$ caused by action $\hat{a}^i_{\tau+1}$. Specifically, task-relevant context triplets are retrieved from the current KG $\hat{\mathcal{G}}_{\tau}^i$ and used as state information, which is the same process as the retrieval $\mathcal{R}(\hat{\mathcal{G}}^i_{\tau},\eta)$ described in the fast-thinking section. Then with these triplets and $\hat{a}^i_{\tau+1}$ as input, LLM is prompted to predict the next observation $\hat{o}_{\tau+1}^i$. This can be formally written as $\hat{o}_{\tau+1}=LLM(P^{\text{obs}};\mathcal{R}(\hat{\mathcal{G}}^i_{\tau},\eta),\hat{a}_{\tau+1}^i)$. In this prediction, the LLM is asked to infer plausible future observations using its commonsense knowledge and reasoning capabilities. 

Then, $\hat{o}_{\tau+1}^i$ is parsed into triplets which are incorporated into $\hat{\mathcal{G}}_{\tau}^i$ to produce the next latent state (KG) $\hat{\mathcal{G}}_{\tau+1}^i$, following the process of updating KG introduced in the fast-thinking section, i.e. $\hat{\mathcal{G}}_{\tau+1}^i=\text{Update}(\hat{\mathcal{G}}_{\tau}^i,\hat{o}_{\tau+1}^i,\hat{a}_{\tau+1})$. The newly acquired triplets in updating process are denoted as $\tau_{\text{new}}$.

\noindent
{\bf Weight Update.} As shown in the blue box of Figure \ref{fig:tsmc_planning}, each particle $i$ is assigned a new weight $\omega_{\tau+1}^i$ which is updated as $\omega_{\tau+1}^i\leftarrow\exp(\lambda\delta^i_{\tau+1})$. Specifically, $\delta^i_{\tau+1}$ is the task progress detector and is the LLM evaluation on whether the transition $(\hat{\mathcal{G}}_{\tau}^i, \hat{a}_{\tau}^i, \hat{\mathcal{G}}_{\tau+1}^i)$ makes progress toward task completion or not. Formally, we can write $\delta^i_{\tau+1}=LLM(P^{\text{prog}}; \hat{a}_{\tau}^i, \mathcal{R}(\hat{\mathcal{G}}^i_{\tau}),  \tau_{\text{new}}, \eta)$, where $P^{\text{prog}}$ is the prompt template in the task progress evaluation, $\delta^i_{\tau+1}\in\{0, 1\}$, and $\tau_{\text{new}}$ is obtained in the propagation process. In contrast to previous LLM-based evaluation methods \citep{zheng2025survey}, in our approach, the binary response of LLM and concise representation provided by context triplets can reduce hallucination and make LLM focus on task-related information in the input.

\noindent
{\bf Resampling.} The resampling step is to probabilistically keep particles with good task progress and discard particles without poor progress. At planning step $\tau$, the likelihood of a particle $i$ being kept is proportional to its weight $\omega_{\tau}^i$.

Repeating the operations of propagation, weight update and resampling over the planning horizon $H$, the remaining particles are close to successful trajectories with a large probability. This can be guaranteed by the theoretical properties of TSMC \citep{del2006sequential}. After planning step $H$, uniformly sample a particle $j\sim\text{Unif}([1,\ldots,N])$, and apply the first action in the particle $j$ (i.e. $\hat{a}_1^j$) to the actual environment, as shown in Figure \ref{fig:proposed_framework}. The theoretical justification is in Appendix \ref{sec:app_slow_thinking}.

\section{Experiments}
\label{sec:exp}
\subsection{Experimental Settings}
\label{sec:setting}
\noindent
{\bf Benchmarks.} We benchmarked NeSyFS on three widely used text-based environments: ALFWorld \citep{shridhar2021alfworld}, WebShop \citep{yao2022webshop}, and ScienceWorld \citep{wang2022scienceworld}. ALFWorld evaluates embodied agents on household tasks, WebShop \citep{yao2022webshop} emulates multi-step decision-making tasks in an online shopping website environment, and ScienceWorld \citep{wang2022scienceworld} assesses procedural and scientific reasoning in educational scenarios. For evaluation, ALFWorld uses binary task success, while WebShop and ScienceWorld provide dense reward signals, enabling evaluation based on both success rate and average reward, calculated as the mean reward across all tasks. Additional benchmark details are presented in Appendix \ref{sec:app_benchmarks}.

\begin{table*}[ht]
    \centering
    \caption{Performance comparison of NeSyFS with ReAct, Reflexion, RAFA and SwiftSage across ALFWorld, WebShop and ScienceWorld. SR and AR denote success ratio and average reward, respectively.}
    \renewcommand{\arraystretch}{1.}
    \begin{tabular}{l|l|c|c|c|c|c|c}
    \hline
    \multirow{2}{*}{\textbf{Model}} & \multirow{2}{*}{\textbf{Method}} & {\textbf{ALFWorld}} & \multicolumn{2}{c}{\textbf{WebShop}} & \multicolumn{2}{c}{\textbf{ScienceWorld}} &  \multirow{2}{*}{\textbf{Average}}\\
      &  & SR & AR & SR & AR & SR &   \\\hline\hline
    \multirow{6}{*}{GPT-5-mini}  & ReAct & $71.2$ & $42.5$ & $32.7$ & $65.1$ & $28.3$ & $44.1$ \\
      & Reflexion &  $75.6$ & $49.5$ & $35.7$ & $69.7$ & $31.5$ & $47.6$ \\
      & ABBEL &  $72.1$ & $43.2$ & $33.3$ & $65.9$ & $29.5$ & $44.9$ \\
      & RAFA &  $73.5$ & $40.1$ & $27.3$ & $70.2$ & $34.5$ & $45.1$ \\
      & SwiftSage & $76.7$ & $52.3$ & $38.2$ & $72.1$ & $35.7$ & $51.0$ \\
      & \textbf{NeSyFS} & $\bm{91.1}$ & $\bm{61.2}$ & $\bm{51.3}$ & $\bm{82.2}$ & $\bm{61.2}$ & $\bm{63.5}$ \\\hline
    \multirow{6}{*}{GPT-5}  & ReAct & $78.2$ & $49.6$ & $38.7$ & $72.2$ & $35.5$ & $50.8$ \\
      & Reflexion &  $80.3$ & $52.3$ & $40.2$ & $75.7$ & $38.9$ & $53.1$ \\
      & ABBEL &  $77.6$ & $51.2$ & $39.8$ & $73.9$ & $36.0$ & $51.1$ \\
      & RAFA &  $83.2$ & $48.1$ & $35.3$ & $76.3$ & $42.7$ & $53.7$ \\
      & SwiftSage & $81.9$ & $56.3$ & $41.2$ & $75.3$ & $40.2$ & $54.4$ \\
      & \textbf{NeSyFS} & $\bm{93.6}$ & $\bm{69.4}$ & $\bm{63.3}$ & $\bm{86.1}$ & $\bm{69.4}$ & $\bm{75.3}$ \\\hline
    \multirow{6}{*}{Llama-3.3-70B}  & ReAct & $79.6$ & $45.6$ & $37.1$ & $69.2$ & $33.1$ & $49.9$ \\
      & Reflexion & $81.2$ & $47.3$ & $38.9$ & $73.1$ & $36.9$ & $52.3$ \\
      & ABBEL &  $78.6$ & $45.1$ & $36.9$ & $70.1$ & $32.5$ & $49.3$ \\
      & RAFA & $82.3$ & $47.2$ & $37.1$ & $75.1$ & $39.2$ & $52.9$ \\
      & SwiftSage & $80.6$ & $51.2$ & $39.9$ & $76.1$ & $41.3$ & $53.9$ \\
      & \textbf{NeSyFS} & $\bm{89.6}$ & $\bm{68.1}$ & $\bm{62.5}$ & $\bm{85.2}$ & $\bm{67.1}$ & $\bm{73.0}$ \\\hline
    \end{tabular}
    \label{tab:table_full_result}
\end{table*}

\noindent
{\bf Agent Models.} In the evaluations, we use GPT-5, GPT-5-mini \citep{singh2025openai} and Llama-3.3-70B-Instruct \cite{grattafiori2024llama} as the underlying models. GPT-5 variants serve as proprietary models, while Llama-3.3 represents the open-source counterpart.

In the following, we first conduct experiments to compare the overall framework against some previous representative methods. Then, the effect of reflection module is investigated empirically, independently of the slow-thinking module. Finally, the performance of planning algorithm in the slow-thinking module is specifically evaluated in Appendix by removing the reflection module. Every result reported here is an average of three random seeds.

\subsection{Overall Evaluation}
\label{sec:comparison}

\noindent
{\bf Baseline.}
Some representative decision making methods of LLM agent are selected as baselines. Details of each baseline are introduced in Appendix \ref{sec:app_baselines}.

\begin{itemize}
    \item ReAct \citep{yao2022react}: This method reasons about next action by using CoT \cite{wei2022chain}, laying the foundation of LLM decision making.
    \item Reflexion \citep{shinn2023reflexion}: This method reflects every failed trajectory and helps LLMs improve after each failed attempt. 
    \item ABBEL \citep{lidayan2025abbel}: It uses LLM to summarize observation and full interaction history to infer the belief state.
    \item RAFA \citep{liu2023reason}: At each time step, the agent first conducts a short-horizon planning through tree search, executes the first action of the plan, and then replans at next step. 
    \item SwiftSage \citep{lin2023swiftsage}: In this fast-slow thinking framework, the Swift module represents fast and intuitive thinking, and the Sage module plans and grounds subgoals to emulate deliberate thought processes.  
\end{itemize}

\noindent
{\bf Results.} Table \ref{tab:table_full_result} presents the performance comparison between NeSyFS and several representative baselines, including ReAct, Reflexion, ABBEL, RAFA, and SwiftSage, across ALFWorld, WebShop, and ScienceWorld. The results show that NeSyFS consistently and significantly outperforms all baseline methods. In particular, NeSyFS achieves nearly a 100\% improvement over ReAct and Reflexion, which serve as the foundation of many state-of-the-art LLM agents, highlighting the effectiveness of the proposed KG-based memory, reflection, and slow-thinking modules.

Furthermore, the substantial improvement over summarization-based method, such as ABBEL, suggests that KG-based contextual representations provide more effective state representations for decision making than summarization-based belief states, since the summarization of history may lose critical information for belief updates. RAFA performs short-horizon planning at each time step through tree search, a mechanism conceptually related to our slow-thinking module. The superior performance of NeSyFS over RAFA therefore underscores the importance of integrating KG-based memory and reflection mechanisms in addition to planning. SwiftSage represents a typical fast–slow thinking framework, and the advantage of NeSyFS further demonstrates the benefits of KG-based memory in enhancing the synergy between fast and slow reasoning processes. Comparisons with previous KG-augmented LLM-agent methods \citep{agarwal2025l3m+,anokhin2025arigraph} are provided separately in the next section.

\subsection{KG-provided Context and Reflection}
\label{sec:reflection_result}
In this section, we first evaluate the correctness of the KG-augmented reflection module, and then demonstrate the advantage and effect of KG-retrieved context triplets in the overall task-completion evaluations.

The reflection module of NeSyFS leverages KG-retrieved triplets as contextual information. In the first part of experiment, to evaluate the effect of KG-provided context in reflection, we compare the correctness of our reflection module against baselines that use the entire interaction history and LLM-summarized belief state as context, which are short as "history" and "belief", respectively. The proposed reflection method is short as "KG". All the methods are evaluated on a set of randomly-collected trajectories in which the correctness of each action is manually annotated.

We adopt two evaluation metrics. The first is Total Detection Errors (TDE), defined as the sum of false positives (misaligned actions that are not detected) and false negatives (correct actions incorrectly classified as misaligned). TDE measures the overall number of reflection errors. The second metric is Effective Reliability (ER), defined as $\frac{TP-FP}{TP+FP}$, where $TP$ and $FP$ denote the numbers of true positives and false positives, respectively. Hence, ER evaluates the reliability of actions that are approved by the reflection module.

\begin{table*}[ht]
    \centering
    \caption{Comparison of reflection with different representations of context.}
    \label{tab:table_tde_er}
    \renewcommand{\arraystretch}{1.}
    \begin{tabular}{llcc|cc|cc}
    \hline
      \multirow{2}{*}{\textbf{Model}} & \multirow{2}{*}{\textbf{Method}} & \multicolumn{2}{c}{\textbf{ALFWorld}} & \multicolumn{2}{c}{\textbf{WebShop}} & \multicolumn{2}{c}{\textbf{ScienceWorld}} \\
       & & TDE & ER & TDE & ER & TDE & ER \\\hline
       \multirow{3}{*}{{GPT-5}} & History  & 57 & {0.55} & 87 & 0.51 & 95 & 0.53 \\
          & Belief  & 52 & {0.59} & 89 & 0.50 & 89 & 0.55 \\
        & \textbf{KG}  & \textbf{42} & \textbf{0.71} & \textbf{61} & \textbf{0.66} & \textbf{77} & \textbf{0.65} \\\hline
       \multirow{3}{*}{{GPT-5-mini}} & History  & 82 & {0.45} & 113 & 0.43 & 135 & 0.42 \\
          & Belief  & 77 & {0.41} & 110 & 0.42 & 125 & 0.45 \\
        & \textbf{KG}  & \textbf{58} & \textbf{0.65} & \textbf{72} & \textbf{0.58} & \textbf{91} & \textbf{0.59} \\\hline
    \end{tabular}
\end{table*}

The performance comparison is shown in Table \ref{tab:table_tde_er}. We can see that the reflection method using KG-provided context significantly outperforms that of using interaction history or LLM-summarized belief state as context. This is because KG-retrieved triplets can provide concise state representation, making the LLM focus on decision-making-related information in the prompt when reflecting. Notably, the ER of "KG" method is significantly higher than baselines, meaning that KG-provided context could prevent more wrong actions from being applied into the environment. "History" performs worse since the noisy and redundant information in the interaction history can distract LLM. The problem of "Belief" is that the LLM-based summarization can lose important information.

\begin{figure}
    \centering
    \subfigure[GPT-5-mini]{
        \includegraphics[width=3in]{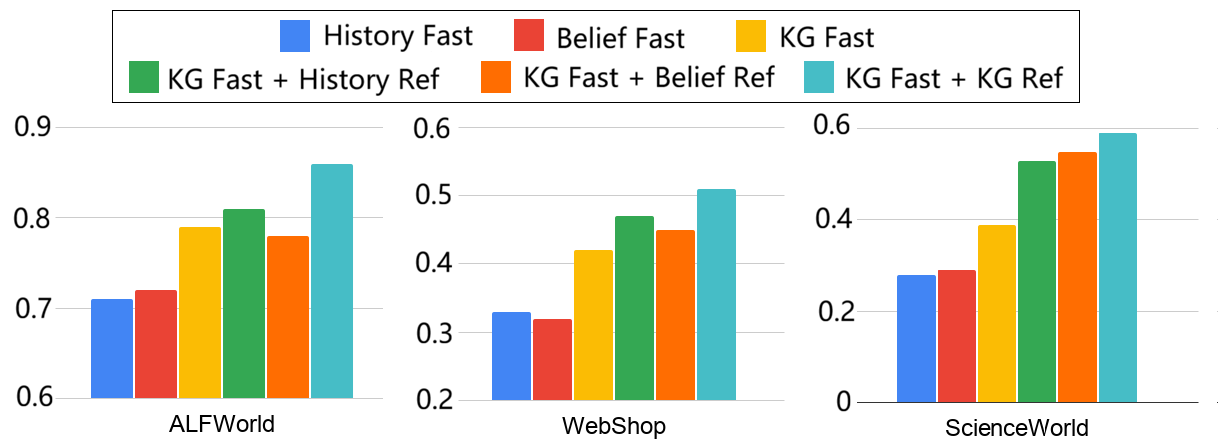}    
    }
    \subfigure[GPT-5]{
        \includegraphics[width=3in]{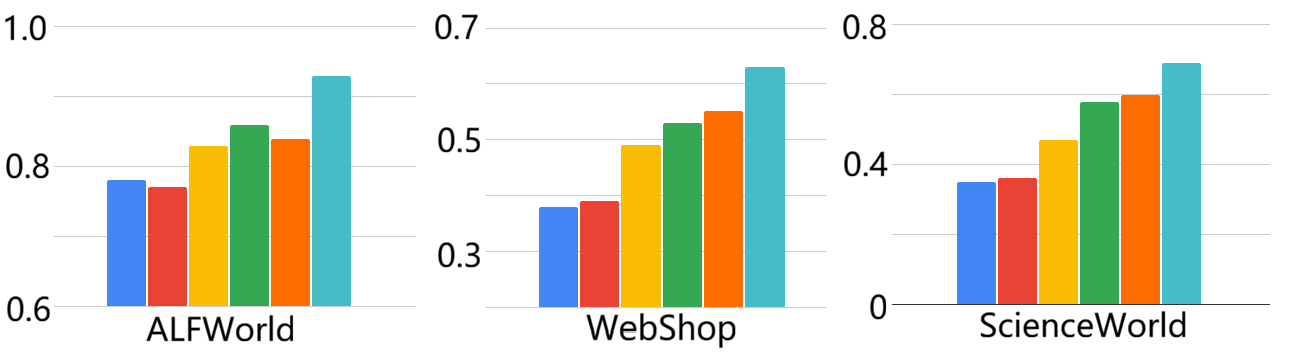}    
    }
    \caption{Performance comparison of different reflection methods.}
    \label{fig:reflection_full}
\end{figure}

In the second part of the experiment, we empirically investigate the effect of KG-provided context within the fast-thinking and reflection modules by comparing different methods of forming the context to approximate the underlying environment state. Both GPT-5 and GPT-5-mini are used as the underlying LLMs in these evaluations. The evaluation metric, corresponding to y-axis in the plots, is the task success rate.

The methods “History Fast”, “Belief Fast”, and “KG Fast” use the full interaction history, an LLM-summarized belief state, and KG-retrieved triplets, respectively, as context for action selection in the fast-thinking module, without employing any reflection module or slow-thinking module. In contrast, “History Ref”, “Belief Ref”, and “KG Ref” incorporate the reflection module while using the same three forms of context representations. Specifically, in the fast-thinking and reflection modules, the KG is retrieved and updated using the same procedures as those employed in NeSyFS, as described in Section \ref{sec:fast}. Besides, the belief state is updated using the same method as ABBEL introduced in Appendix \ref{sec:app_baselines}. No slow-thinking module is used in any of these evaluations here. Notably, the “Fast” method are conceptually equivalent to previous KG-augmented LLM agents \citep{agarwal2025l3m+,anokhin2025arigraph}.

As illustrated in Figure \ref{fig:reflection_full}, incorporating the reflection module consistently improves performance across all benchmarks. Moreover, the use of KG-provided context leads to additional performance gains, demonstrating both the importance of reflection in mitigating misalignment with task objectives and the effectiveness of the symbolic, structured representations provided by the KG.

\section{Conclusion}
In this work, we propose a neuro-symbolic framework which addresses the challenges introduced by partial observability in a unified manner. Under partial observability, the agent may have difficulties on latent state inference, task objective misalignment, and planning under uncertainty. In NeSyFS, the latent state of the environment is represented and maintained as a memory KG, and triplets retrieved from KG are used as context in every module of NeSyFS. In addition, a KG-augmented reflection module is proposed to address the misalignment of the task objective. Besides, a neuro-symbolic TSMC-style planning algorithm is proposed to tackle the uncertainty in observation prediction and task progress evaluation. We empirically validate the proposed framework and its modules in three popular benchmarks.

\bibliography{main}

\newpage
\appendix



\section{Algorithm Description}
\label{sec:algos}
\begin{algorithm}
\caption{Neuro-symbolic Fast-slow Thinking Framework for LLM Agent}
\label{alg:nesyfs}
\begin{algorithmic}[1]
\Require Task description $\eta$; memory KG $\mathcal{G}$; number of reflections for one fast-thinking action $K$; prompts for fast-thinking action generation $P^{\text{fast}}$ and reflection $P^{\text{ref}}$; operations of updating KG $\text{Update}(\mathcal{G},o,a)$ and retrieval from KG $\mathcal{R}(\mathcal{G}, \eta)$; planning algorithm in the slow-thinking module $\text{TSMC-plan}(\eta,\mathcal{G},o)$, described in Algorithm \ref{alg:tsmc-plan};
\State Initialize the memory KG $\mathcal{G}_0\leftarrow\emptyset$;
\State Initialize the environment and get initial observation $o_0$;
\For{$t=1,\ldots,H$}
\State $a_t\sim\text{LLM}(P^{\text{fast}};\mathcal{R}(\mathcal{G}_{t-1},\eta),o_{t-1},\eta)$; \textit{\%Generate fast-thinking action}
\State $k\leftarrow0$;
\While{$k<K$}
\State $\bar{r}, h\sim\text{LLM}(P^{\text{ref}};a_t,o_{t-1},\mathcal{R}(\mathcal{G}_{t-1},\eta),\eta)$; \textit{\%Reflect with thinking step-by-step, where $h$ is the thinking of reflection}
\If{$\bar{r}$ is No}
\State $a_t\sim\text{LLM}(P^{\text{fast}};\mathcal{R}(\mathcal{G}_{t-1},\eta),o_{t-1},\eta,h)$; \textit{\%Generate fast-thinking action with reflection thinking}
\State $k\leftarrow k+1$;
\Else
\State Break;
\EndIf
\EndWhile
\If{$k\ge K$}
\State $\hat{\mathcal{G}}\leftarrow\mathcal{G}_{t-1}$; \textit{\%Copy current KG}
\State $a_t\sim\text{TSMC-plan}(\eta,\hat{\mathcal{G}},a_t,o_{t-1})$; \textit{Initiate the planning algorithm in the slow-thinking module}
\EndIf
\State Apply $a_t$ into the environment, and obtain new observation $o_t$ and reward $r_t$;
\State $\mathcal{G}_t\leftarrow\text{Update}(\mathcal{G}_{t-1},o_t,a_t)$;
\EndFor
\end{algorithmic}
\end{algorithm}

The algorithm table of the proposed NeSyFS framework is presented in Algorithm \ref{alg:nesyfs}. In every iteration, the fast-thinking action is first generated in line 4. Then, from line 5 to 14, the reflection is conducted to ensure the alignment of action and task objective. If reflection fails repeatedly, the slow-thinking module is initiated in line 16, where the first action of the best particle is produced as output. In line 18, the selected action $a_t$ is applied into the environment. In line 19, the memory KG is updated with action $a_t$ and new observation $o_t$. In every experiment, if not stated specifically, $K$ is selected to be $3$.

\begin{algorithm}[ht]
\caption{TSMC-plan($\eta,\mathcal{G}_0,a_0,o_0$)}
\label{alg:tsmc-plan}
\begin{algorithmic}[1]
\Require Task description $\eta$; initial memory KG $\mathcal{G}_0$; initial observation $o_0$; number of particles $N$; planning horizon $L$; prompts for fast-thinking $P^{\text{fast}}$, observation prediction $P^{\text{obs}}$ and task progress detection $P^{\text{prog}}$ in the slow-thinking module; Operation of updating KG $\text{Update}(\mathcal{G},o,a)$ and retrieval from KG $\mathcal{R}(\mathcal{G},\eta)$;
\State Initialize $\hat{o}^i_1\leftarrow o_0, \hat{\mathcal{G}}_1^i\leftarrow\mathcal{G}_0, \hat{a}^i_1\leftarrow a_0$, for $i=1,\ldots,N$;
\For{$\tau=2,\ldots,L$}
\For{$i=1,\ldots,N$}
\State $\hat{a}^i_{\tau}\leftarrow\text{LLM}(P^{\text{fast}};\mathcal{R}(\hat{\mathcal{G}}^i_{\tau-1},\eta),\hat{o}^i_{\tau-1},\eta)$;
\State $\hat{o}^i_{\tau}\leftarrow \text{LLM}(P^{\text{obs}};\mathcal{R}(\hat{\mathcal{G}}^i_{\tau-1},\eta),\hat{a}_{\tau}^i)$;
\State $\hat{\mathcal{G}}_{\tau}^i\leftarrow \text{Update}(\hat{\mathcal{G}}_{\tau-1}^i,\hat{o}_{\tau}^i,\hat{a}_{\tau}^i)$, where the newly acquired triplets are denoted as $\nu_{\text{new}}^i$;
\State $\delta^i_{\tau}\leftarrow \text{LLM}(P^{\text{prog}}; \hat{a}_{\tau-1}^i, \mathcal{R}(\hat{\mathcal{G}}^i_{\tau-1}, \eta), \nu^i_{\text{new}}, \eta)$;
\State $\omega_{\tau}^i\leftarrow \exp(\lambda\delta_{\tau}^i)$;
\EndFor
\State Define new particles $(\{\tilde{a}^i_{\iota}\}_{\iota=1}^{\tau},\tilde{o}^i,\tilde{\mathcal{G}}^i,\tilde{\delta}^i,\tilde{\omega}^i)\leftarrow(\{\hat{a}^i_{\iota}\}_{\iota=1}^{\tau},\hat{o}^i_{\tau},\hat{\mathcal{G}}^i_{\tau},{\delta}^i_{\tau},{\omega}^i_{\tau})$, for $i=1,\ldots,N$;
\For{$i=1,\ldots,N$}
\State Sample $j\sim\text{Categorical}(\tilde{\omega}^1,\ldots,\tilde{\omega}^N)$;
\State $(\{\tilde{a}^i_{\iota}\}_{\iota=1}^{\tau},\hat{o}^i_{\tau},\hat{\mathcal{G}}^i_{\tau},{\delta}^i_{\tau},{\omega}^i_{\tau})\leftarrow (\{\tilde{a}^j_{\iota}\}_{\iota=1}^{\tau},\tilde{o}^j,\tilde{\mathcal{G}}^j,\tilde{\delta}^j,\tilde{\omega}^j)$;
\EndFor
\EndFor
\State Sample $l\sim\text{Uniform}(1,\ldots,N)$;
\State \textbf{return} $\hat{a}_2^l$;
\end{algorithmic}
\end{algorithm}

The proposed planning algorithm in the slow-thinking module is presented in Algorithm \ref{alg:tsmc-plan}. The operations for propagation of every particle is conducted in lines 4, 5 and 6. The weight update of every particle is from line 7 to 8. Resampling is from line 10 to 14, where the historical actions of every particle are kept in the particle replacement. Finally, in line 17, the first action $\hat{a}^l_2$, not the initial one $a_0$, is returned. In every experiment, the number of particles $N$ is set to 6, and the planning horizon $L$ is 5. Because the observation prediction may not be accurate and prediction mistakes can be accumulated along the time axis, $L$ cannot be too large. In order to control the token consumption, $N$ cannot be too large as well. The temperature parameter $\lambda$ is selected to 1.

\section{Related Work}
\label{sec:related_work}

In traditional RL, partial observability is addressed by estimating the belief state \citep{rodriguez1999reinforcement,roy2005finding,li2009multi}, maintaining an estimate of the latent environment state. However, such approaches typically rely on accurate knowledge or estimate of transition or observation models \citep{xiang2021recent}, which are difficult to satisfy in complex and heterogeneous agentic settings. In LLM, many previous works on LLM agents address partial observability by treating interaction history as a surrogate for the latent state \citep{yao2022react,ma2024agentboard,xi2024agentgym,shinn2023reflexion,zhao2024expel,lei2025large}. However, because agentic data are typically collected under outcome-level supervision—where a single terminal success signal is assigned to an entire episode—such histories frequently contain weakly informative or task-irrelevant content. Consequently, conditioning the decision making of LLM on full trajectories entangles state-relevant signals with spurious details \citep{chung2025evaluating}, incurs substantial computational overhead \citep{kang2025acon}, and ultimately degrades performance \citep{lei2025large}. 

Some previous work on the LLM agent has leveraged recent advances in context management to address partial observability and has largely focused on the summarization and retrieval of interaction histories, including compression-based summarization and segmented memory retrieval \citep{lu2025scaling,kang2025acon,chen2025iterresearch,lidayan2025abbel,pan2026secom}. While effective for controlling context length, these approaches implicitly assume that the interaction history—whether summarized or selectively retrieved—constitutes an adequate representation of the underlying state. This assumption introduces additional computational overhead and renders agents susceptible to information loss. Besides, unstructured memory (e.g., full history, compression-based summarization) stores information as disconnected text fragments, making it difficult to retrieve logically related facts. In contrast, our framework uses structured memory, organizing knowledge as a graph. A recent work \citep{agarwal2025l3m+} uses an external knowledge graph to represent the latent state for LLM agent. However, this work assumes that domain knowledge of the environment, i.e. PDDL, is provided, and it does not challenges of reconciling knowledge gap and planning under uncertainty which are introduced by partial observability.

In addition, many agentic approaches prioritize aligning LLMs for effective interaction rather than learning compact state representations from histories. For instance, \citep{qin2023toolllm,zhang2024agentohana} unify instruction-following pipelines, while methods such as Exploration-based Trajectory Optimization \citep{song2024trial} and Group-in-Group Policy Optimization \citep{feng2025group} optimize behavior using trajectory-level or group-level feedback. AgentEvol \citep{xi2024agentgym} incorporates RL rewards into supervised fine-tuning (SFT), and \citep{chen2025atlas} demonstrates gains from curated interaction subsets. However, these approaches still uniformly treat the full action–observation history as the state representation, and their training or fine-tuning process of LLM is expensive and can incur significant computational overhead. In contrast, our framework does not change any parameters of pre-trained LLMs.

\section{Supplementary Materials to Methodology}
\label{sec:app_method}
In this section, we present more details about the proposed NeSyFS framework.

\begin{figure}
    \centering
    \includegraphics[width=3in]{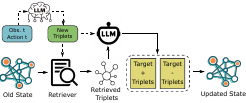}
    \caption{Diagram of updating KG with new observation $o_t$ and action $a_t$.}
    \label{fig:thrust1_kg_target}
\end{figure}

\subsection{Fast-thinking Module}
\label{sec:app_fast}
\subsubsection{Update to KG}
\label{sec:update_KG}
The diagram of updating KG $\mathcal{G}_t$ with new observation $o_t$ and action $a_t$ is shown in Figure \ref{fig:thrust1_kg_target}, where $o_t$ is the response of the environment to $a_t$. Given a new observation $o_t$, the LLM agent first parses $o_t$ into a set of entities $v_t$ and triplets $\tau_t$ describing attributes and relations of $v_t$. For example, an observation may generate triplets such as $(\text{apple 1}, \text{at}, \text{sidetable 1})$ or $(\text{you}, \text{cleaned}, \text{tomato 1})$. The extracted triplets are then filtered by removing those already contained in the KG $\mathcal{G}_t$, resulting in a set of new triplets $\tau_{\text{new}}$ extracted from $o_t$. To identify previously stored knowledge related to the entities in $v_t$, the agent retrieves from $\mathcal{G}_t$ all triplets connected to these entities, denoted as the set of existing triplets $\tau_{\text{old}}$. The LLM is subsequently prompted to compare $\tau_{\text{new}}$ and $\tau_{\text{old}}$ in order to determine which triplets in $\tau_{\text{old}}$ have become outdated and should be replaced by corresponding triplets in $\tau_{\text{new}}$. For example, the triplet $(\text{you}, \text{at}, \text{cabinet 1})$ in $\tau_{\text{old}}$ may be replaced by $(\text{you}, \text{at}, \text{fridge 1})$ in $\tau_{\text{new}}$ to reflect a change in the entity’s location. Finally, according to the LLM’s output, outdated triplets are removed from $\mathcal{G}_t$ and replaced with updated triplets in $\tau_{\text{new}}$, thereby updating $\mathcal{G}_t$ with $o_t$ and producing KG $\mathcal{G}_{t+1}$ for next time step.

\subsubsection{Retrieval from KG}
\label{sec:retrieval_KG}
Given the task description $\eta$ (e.g., “put a cool egg into a microwave”), rather than providing the entire KG $\mathcal{G}_t$ to LLM, our approach first retrieves task-relevant triplets $\tau_t$ from $\mathcal{G}_t$ and uses $\tau_t$ as context for decision making.
Specifically, by prompting LLM, the last observation $o_t$ is first parsed into entities $v_q$ and relations $E_q$ which are related with solving the task $\eta$. Subsequently, all incoming and outgoing triplets associated with each entity in $V_q$ are retrieved from the present KG $\mathcal{G}_t$ up to a predefined maximum depth, and retrieved triplets are denoted as $\tau_t$. 
The triplets $\tau_t$ are then supplied to the LLM as structured context for reactive decision making in fast thinking. In addition, the KG and retrieval method introduced above are also used in the slow thinking and reflection modules. This KG retrieval method is denoted as $\mathcal{R}(\cdot,\cdot)$.


\subsection{Slow-thinking Module}
\label{sec:app_slow_thinking}
Denote a trajectory as $\bm{\xi}_T=\{\bm{x}_t\}_{t=1}^T$ where $\bm{x}_t=(o_t,\mathcal{G}_t,a_t)$, representing observation, latent state (KG), and action at time step $t$, respectively. The target distribution $\sigma(\bm{\xi}_T)$ is defined as the distribution of trajectories that complete the task successfully. 

\paragraph{Theoretical Justification.} Unlike standard sequential Monte Carlo (SMC) methods \citep{del2006sequential}, we employ twisted SMC (TSMC), which introduces a sequence of twist functions $\{\psi_{\tau}\}_{\tau=1}^{T-1}$ to bias intermediate particle distributions toward high-probability regions of the target distribution $\sigma(\bm{\xi}_T)$. For a general proposal distribution q, prior work \citep{zhao2024probabilistic,feng2025step} defines the incremental importance weight as
\begin{eqnarray}
   \small
   \omega_{\tau}^k &\leftarrow& \frac{ p_0(s_{\tau}^k \mid s^k_{1:\tau-1}) }{ q(s_{\tau}^k \mid s^k_{1:\tau-1}) } \frac{ \psi_{\tau}(s_{1:\tau}^k) }{ \psi_{\tau-1}(s_{1:\tau-1}^k) }, \quad \tau < T; \nonumber \\
    \omega_{\tau}^k &\leftarrow& \frac{ p_0(s_{\tau}^k \mid s^k_{1:\tau-1}) }{ q(s_{\tau}^k \mid s^k_{1:\tau-1}) } \frac{ \phi(s_{1:\tau}^k) }{ \psi_{\tau-1}(s_{1:\tau-1}^k) }, \quad \tau = T 
   \label{omega_org}
\end{eqnarray}
Here, $p_0$ and $q$ denote, respectively, the base-model distribution and the proposal distribution conditioned on the current history. The twist function $\psi_{\tau}(\bm{x}_{1:\tau})$ is designed so that the twisted intermediate distribution better approximates the corresponding target marginal $\sigma(\bm{x}_{1:\tau})$. In particular, the optimal twist is proportional to the expected terminal potential under future rollouts of the base model and can therefore be interpreted as a value-like quantity measuring the expected future utility of a partial trajectory \citep{zhao2024probabilistic}. Specifically,
\begin{equation}
\psi_{\tau}^{*}(\bm{x}_{1:\tau})\propto\mathbb{E}_{p_0}\left[\phi(\bm{x}_{1:T})\mid\bm{x}_{1:\tau}\right].
\end{equation}
This interpretation provides the theoretical basis for relating the twist ratio to task progress in our planning formulation. Specifically, consider particle $i$ and the transition $\left(\hat{\mathcal{G}}_{\tau}^{i},\hat{a}_{\tau}^{i},\hat{\mathcal{G}}_{\tau+1}^{i}\right)$.
We use $\delta_{\tau+1}^{i}$ in Algorithm \ref{alg:tsmc-plan} to indicate whether this transition makes measurable progress toward task completion. Since the ratio $\frac{\psi_{\tau+1}(\bm{x}_{1:\tau+1}^{i})}{\psi_{\tau}(\bm{x}_{1:\tau}^{i})}$ measures the relative increase in the estimated future value of the particle after the transition, it is natural to parameterize this multiplicative change as
\begin{equation}
\frac{ \psi_{\tau+1}(\bm{x}_{1:\tau+1}^{i}) }{ \psi_{\tau}(\bm{x}_{1:\tau}^{i}) } \approx \exp\left( \lambda \delta_{\tau+1}^{i} \right), \qquad \lambda > 0 . \label{eq:progress-twist-ratio}
\end{equation}
Under this parameterization, when no progress is detected, $\delta_{\tau+1}^{i} = 0$, we obtain $\exp\left( \lambda \delta_{\tau+1}^{i} \right) = \exp(0) = 1$, so that the incremental weight remains unchanged. In contrast, when progress is detected, $\delta_{\tau+1}^{i} > 0$, we have $\exp\left( \lambda \delta_{\tau+1}^{i} \right) > 1$, thereby assigning greater weight to particles whose transitions advance the task. The coefficient $\lambda$ therefore controls the strength of the progress-based preference. 

In our implementation, the same pretrained LLM, e.g., GPT-5, is used both as the base model and as the proposal model. This corresponds to the "base model as proposal" setting of \citep{zhao2024probabilistic}, for which $q(s_{\tau}^k \mid s^k_{1:\tau-1}) = p_0(s_{\tau}^k \mid s^k_{1:\tau-1})$. Consequently, the proposal-correction factor in  \eqref{omega_org} cancels. The incremental weight therefore simplifies to
\begin{eqnarray}
\omega_{\tau}^k &\leftarrow& \frac{ \psi_{\tau}(s_{1:\tau}^k) }{ \psi_{\tau-1}(s_{1:\tau-1}^k) }, \quad \tau < T \nonumber \\
\omega_{\tau}^k &\leftarrow& \frac{ \phi(s_{1:\tau}^k) }{ \psi_{\tau-1}(s_{1:\tau-1}^k) }, \quad \tau = T  \label{tsmc-new-weight-update}
\end{eqnarray}
Combining \eqref{tsmc-new-weight-update} with the progress-based approximation in  \eqref{eq:progress-twist-ratio}, the incremental weight for particle $i$ can be written as $\omega_{\tau+1}^{i} \approx \exp\left( \lambda \delta_{\tau+1}^{i} \right)$. This establishes the connection between the progress-based weighting rule in Algorithm \ref{alg:tsmc-plan} and the incremental importance weights of TSMC \citep{del2006sequential,feng2025step}.

\section{Experiments}
\label{sec:app_exp}
\subsection{Benchmarks}
\label{sec:app_benchmarks}
The detailed introduction of benchmarks is as follows.
\begin{itemize}
    \item {\bf ALFWorld} \cite{shridhar2021alfworld}: ALFWorld is a text-based benchmark derived from the ALFRED dataset, designed to evaluate an agent’s ability to interpret and execute natural-language instructions in interactive, multi-step household environments. By transforming embodied vision-and-language tasks into textual interactions, it enables systematic evaluation of high-level planning, language grounding, and sequential decision-making capabilities within a simulated environment. 
    \item {\bf WebShop} \cite{yao2022webshop}: WebShop is a large-scale simulated e-commerce benchmark developed for studying grounded language agents in realistic web-interaction settings. Given a natural-language shopping instruction specifying product requirements, an agent must navigate diverse webpage types—including search pages, result pages, product pages, and product-detail pages—while performing actions such as generating search queries, selecting products, choosing item options, reading descriptions, backtracking, and ultimately purchasing the desired item.
    \item {\bf ScienceWorld} \cite{wang2022scienceworld}: ScienceWorld is a text-based benchmark designed to evaluate agents on scientific reasoning and procedural task completion. Grounded in standardized K–12 science curricula, it assesses an agent’s ability to follow instructions, conduct virtual experiments, perform causal reasoning, and manipulate objects across domains such as physics, chemistry, and biology. The benchmark emphasizes multi-step interaction, hypothesis testing, and language grounding in dynamic environments.
\end{itemize}

\begin{table*}[ht]
    \centering
    \caption{Comparison of proposed task progress detector and baselines.}
    \label{tab:table_progress_detector}
    \renewcommand{\arraystretch}{1.}
    \begin{tabular}{llcc|cc|cc}
    \hline
      \multirow{2}{*}{\textbf{Model}} & \multirow{2}{*}{\textbf{Method}} & \multicolumn{2}{c}{\textbf{ALFWorld}} & \multicolumn{2}{c}{\textbf{WebShop}} & \multicolumn{2}{c}{\textbf{ScienceWorld}} \\
       & & TDE & ER & TDE & ER & TDE & ER \\\hline
       \multirow{4}{*}{{GPT-5}} & History  & 45 & {0.61} & 97 & 0.55 & 89 & 0.58 \\
        & Belief  & 40 & {0.63} & 87 & 0.57 & 78 & 0.61 \\
        & Fraction  & 89 & {0.51} & 132 & 0.46 & 105 & 0.49 \\
        & \textbf{Proposed}  & \textbf{29} & \textbf{0.73} & \textbf{69} & \textbf{0.62} & \textbf{53} & \textbf{0.72} \\\hline
       \multirow{4}{*}{{GPT-5-mini}} & History  & 63 & {0.56} & 112 & 0.49 & 99 & 0.51 \\
        & Belief  & 54 & {0.52} & 101 & 0.47 & 85 & 0.57 \\
        & Fraction  & 102 & {0.45} & 153 & 0.41 & 124 & 0.47 \\
        & \textbf{Proposed}  & \textbf{36} & \textbf{0.69} & \textbf{87} & \textbf{0.57} & \textbf{62} & \textbf{0.65} \\\hline
    \end{tabular}
\end{table*}

\subsection{Baselines}
\label{sec:app_baselines}
The baselines are introduced as below with details.
\begin{itemize}
    \item ReAct \citep{yao2022react}: This method lays the foundation for many decision-making methods of LLM agent. In ReAct, the agent first reasons about the next action at each time step by using CoT \citep{wei2022chain} and then generates an action. The full interaction history is appended into the prompt for the LLM to infer the latent state. The source code of ReAct is at \url{https://github.com/ysymyth/ReAct}
    \item Reflexion \citep{shinn2023reflexion}: built upon ReAct, this method proposes a multi-round approach enabling the LLM agent to use the history of previously failed trajectories to refine their planning for the next round. At the end of every failed episode, the LLM agent infers the reason why it fails based on the full trajectory, and append the inferred experience into the prompt to improve its future decision making.
    We evaluate Reflexion with 2 shots. The source code is at \url{https://github.com/noahshinn/reflexion}
    \item ABBEL \cite{lidayan2025abbel}: ABBEL is a framework for long-horizon LLM agents that replaces the full interaction history with a compact natural-language “belief state.” At each step, the agent first updates its belief using the latest observation, then selects the next action conditioned only on this belief rather than the entire trajectory. This creates a belief bottleneck that keeps memory usage nearly constant while maintaining interpretability and reasoning ability. Since each update of belief state can lose some information, instead of updating at every time step, we store each observation and update the belief state together with stored observations only when the length of stored observations exceeds a threshold. In its implementation, we modify the ReAct code by using the same prompts as the original paper.
    \item RAFA \citep{liu2023reason}: This method integrates reasoning, planning and acting through an RL perspective. At each step, the agent first conducts planning through tree search, executes the first action of the plan, receives feedback from environment, and then replans at next step. During planning, the LLM uses past trajectories as in-context examples to simulate the environment dynamics and estimate value functions. For every task, only the initial episode is executed. The source code is at \url{https://github.com/agentification/RAFA_code}
    \item SwiftSage \citep{lin2023swiftsage}: This is a popular fast-slow thinking framework consisting of Swift and Sage modules, where the Swift module represents fast and intuitive thinking, and the Sage module plans and grounds subgoals to emulate deliberate thought processes. For the fairness of comparison, we do not fine-tune the LLM in the Swift module and make it the same as that in the Sage module. The source code is at \url{https://github.com/SwiftSage/SwiftSage/tree/science_world}
\end{itemize}

\subsection{Planning Evaluation}
\label{sec:planning_result}
In the slow-thinking module of NeSyFS, we propose a novel uncertainty-aware planning algorithm that follows the high-level framework of twisted sequential Monte Carlo (TSMC). First, we assess the accuracy of the task progress detector $\delta_t^i$, which is used to update particle weights in the planning algorithm. Second, we investigate the effect of the proposed planning method on improving the overall task-completion performance.

\begin{figure}
    \centering
    \subfigure[GPT-5-mini]{
        \centering
        \includegraphics[width=2.9in]{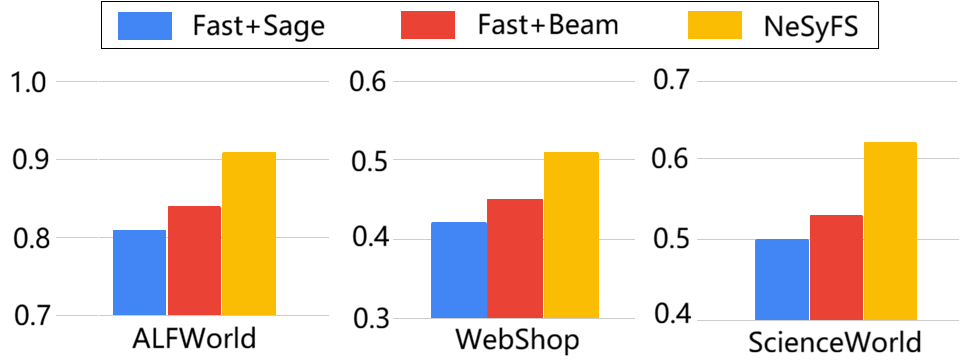}
    }
    \subfigure[GPT-5]{
        \centering
        \includegraphics[width=2.9in]{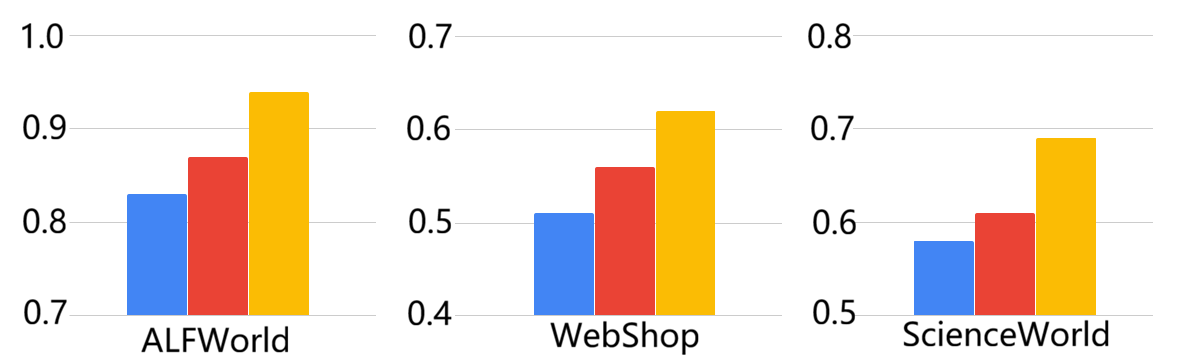}
    }
    \vspace{-10pt}
    \caption{Performance Comparison of fast-slow thinking frameworks with different slow-thinking modules.}
    \label{fig:planning_full_evaluation}
    \vspace{-10pt}
\end{figure}

In the proposed planning algorithm, the particles are filtered by resampling based on weights, so weight update is the key to guide the particles toward the distribution of successful trajectories and find the optimal action plan. Since weight update is based on the task progress detector $\delta_t^i$, in the first part of the experiments, we need to evaluate the accuracy of $\delta_t^i$. Following the previous section, we adopt the total detection error (TDE) and effective reliability (ER) as the evaluation metrics. In contrast, here false positive (FP) refers to transitions that do not make task progress but detected as positive, and false negative (FN) refers to transitions that make task progress but detected as negative. We compare our $\delta_t^i$ with two baselines here. In the first baseline, short as "History", the input state (context) to our detector is replaced with the entire interaction history in text. Same as the ABBEL described above
, the input state in the second baseline, short as "Belief", is the belief state updated by LLM with summary of previous belief state and recent observation. The third baseline, short as "Fraction", is the same as the critics in RAFA \citep{liu2023reason} which prompts LLM to produce an estimate of progress toward task completion, i.e. a fraction between 0 and 1.

For every environment, all the methods are evaluated on a set of randomly pre-collected trajectories, where the ground truth label of making task progress is manually labeled. In "Fraction" method, only the state transition whose critic value is higher than the previous state in the same trajectory is labeled as progressive transition. 
In Table \ref{tab:table_progress_detector}, we demonstrate the comparison of the proposed detector and baselines. It is obvious that the proposed detector outperforms baselines significantly. It performs better than "History", because the KG-retrieved triplets in the proposed method provide concise state representations and can prevent the LLM from being distracted by redundant and noisy information in the raw text of "History" baseline. "Belief" performs better than "History", since it uses LLM to summarize and update belief state with recent observations, but the LLM-based summary in "Belief" may lose some important information for decision making, making "Belief" performs worse than the proposed method. "Fraction" method performs the worst, since the factional output in the estimate of task progress is more difficult for LLM to detect than the binary output. 

In the second part of the experiments, we investigate how much the proposed planning algorithm can improve the overall task-completion performance of the fast-slow thinking framework. We design two baselines by replacing the slow-thinking module of NeSyFS and discard the reflection module, keeping other parts unchanged. First, we use the Sage part of SwiftSage agent \citep{lin2023swiftsage} as the slow-thinking module which reasons about target subgoal to achieve which is then appended as subgoal to the prompt of fast-thinking module, short as "Fast+Sage". Second, we adopt beam search \citep{setlur2025rewarding,freitag2017beam} to realize the slow-thinking module of NeSyFS, short as "Fast+Beam", which expands every sequence in the beam using 3 next actions, computes scores, and selects 2 best sequences with highest score in the expanded sequences to keep in the next iteration with other sequences discarded. The score of every sequence in the beam is the sum of task progress values estimated by $\delta_t^i$. In the "proposed" slow-thinking module, the number of particles is set to $N=6$. In both Beam and proposed method, the planning horizon in the slow thinking is chosen as $H=5$.

The performance comparison of NeSyFS and baselines is shown in Figure \ref{fig:planning_full_evaluation}. These evaluations use GPT-5-mini and GPT-5 as the base LLMs. To rule out the influence of other factors, in every evaluated fast-slow thinking framework, the slow-thinking module is initiated in every 3 steps, and the reflection module is discarded. We can see that the proposed planning algorithm in NeSyFS outperforms than other slow-thinking methods. Fast+Sage performs worst because it simply reasons the next subgoal to achieve in the slow-thinking module, which is not as efficient as the step-by-step planning in beam search and NeSyFS, since setting the next subgoal only does not provide clear guidance on the exact next action to take. Fast+Beam is outperformed by NeSyFS, because the beam search simply selects the best 2 sequences and discards others, which is not robust to the mistakes of LLM-based evaluation on task progress, which can be resolved by the sampling mechanism in the proposed planning method. Note that the margin between NeSyFS and others is larger in WebShop, because the content of website is difficult to predict and hence the prediction of next observation is more difficult in WebShop.

\subsection{Comparison with Tree-search Planning Method}
\label{sec:compare_tree_search}
In this section, we compare the planning algorithm employed in the slow-thinking module of NeSyFS with a conventional tree-search-based planning approach. Specifically, we replace the proposed planning algorithm with the tree-search method introduced in RAFA \cite{liu2023reason}, while keeping all other components of NeSyFS unchanged, thereby constructing a controlled baseline for comparison. We select RAFA because its planning procedure represents a relatively fundamental form of tree-search-based planning, enabling a clear evaluation of the differences between the two approaches. To ensure a fair comparison, the baseline uses the same memory knowledge graph and observation-prediction mechanism as NeSyFS.

For both NeSyFS and the baseline, the maximum number of reflections is set to ($K=1$). In NeSyFS, the planning horizon and number of particles are set to ($L=5$) and ($N=6$), respectively. To ensure a fair comparison, the search breadth and search depth of the tree-search baseline are set to 6 and 5, i.e., ($B=6, U=5$). We evaluate the two methods on all 134 tasks in ALFWorld and the first 100 tasks in WebShop. Performance is assessed using two metrics: token consumption (TC), measured in millions of tokens, and task success rate (SR).

\begin{table}[ht]
    \centering
    \caption{Comparison of proposed and basic tree-search planning method.}
    \label{tab:tree_search_comparison}
    \renewcommand{\arraystretch}{1.}
    \begin{tabular}{ll|cc|cc}
    \hline
      \multirow{2}{*}{\textbf{Model}} & \multirow{2}{*}{\textbf{Method}} & \multicolumn{2}{c}{\textbf{ALFWorld}} & \multicolumn{2}{c}{\textbf{WebShop}} \\
       & & TC & SR & TC & SR \\\hline
       \multirow{2}{*}{{GPT-5}} & Baseline  & 49 & {0.96} & 62 & 0.67 \\
        & \textbf{NeSyFS}  & \textbf{15} & \textbf{0.92} & \textbf{21} & \textbf{0.64} \\\hline
       \multirow{2}{*}{{GPT-5-mini}} & Baseline  & 56 & {0.93} & 69 & 0.58 \\
        & \textbf{NeSyFS}  & \textbf{17} & \textbf{0.90} & \textbf{29} & \textbf{0.53} \\\hline
    \end{tabular}
\end{table}

The comparison results are presented in Table \ref{tab:tree_search_comparison}. The tree-search baseline achieves a modest improvement in task success rate but incurs substantially greater token consumption. This high cost arises because tree-search planning expands multiple successor states of every state at each step, causing the number of evaluated states to grow exponentially with the planning depth. In contrast, the TSMC-style planning method employed by NeSyFS probabilistically eliminates particles associated with limited task progress while maintaining a fixed number of particles at each planning step. Although exhaustive tree search can evaluate a broader range of possible future trajectories and may therefore attain a slightly higher success rate, predicting and assessing these trajectories requires considerable amount of tokens and computational resources. By selectively discarding unpromising particles, the proposed planning algorithm achieves a more favorable trade-off between task-completion performance and computational complexity.

\end{document}